# Parsing Indonesian Sentence into Abstract Meaning Representation using Machine Learning Approach

Adylan Roaffa Ilmy
*School Of Electrical and Informatics Engineering*
*Institut Teknologi Bandung*
Bandung, Indonesia
13516016@std.stei.itb.ac.id

Masayu Leylia Khodra
*School Of Electrical and Informatics Engineering*
*Institut Teknologi Bandung*
Bandung, Indonesia
masayu@informatika.org

*Abstract*—Abstract Meaning Representation (AMR) provides many information of a sentence such as semantic relations, coreferences, and named entity relation in one representation. However, research on AMR parsing for Indonesian sentence is fairly limited. In this paper, we develop a system that aims to parse an Indonesian sentence using a machine learning approach. Based on Zhang et al. work, our system consists of three steps: pair prediction, label prediction, and graph construction. Pair prediction uses dependency parsing component to get the edges between the words for the AMR. The result of pair prediction is passed to the label prediction process which used a supervised learning algorithm to predict the label between the edges of the AMR. We used simple sentence dataset that is gathered from articles and news article sentences. Our model achieved the SMATCH score of 0.820 for simple sentence test data.

## I. Introduction

Banarescu et al. [1] proposed Abstract Meaning Representation (AMR) as a representation that stores many concepts such as semantic relations, coreferences, and named entity relations in a sentence. This representation was designed to determine relation among words called arguments using English Propbank framesets. AMR is a robust semantic representation that can store many semantic concepts in a sentence condensed into one graph, rather than doing each of the task (e.g. coreference resolution, named entity detection) one-by-one. Since different sentences with the same meaning will be represented in the same AMR, structured information can be gathered from different sources that corresponds to the same meaning, easing the task of semantic collections among those different sources [1]. AMR applications can be seen in sentence semantic-similarity based task such as paraphrase detection [2] and multi document summarization [3,4].

The current state-of-the-art AMR parsing system was developed by Zhang et al. [5] that used deep learning approach to parse English to its AMR form. It achieved SMATCH score of 76.3% on LDC2017T10 dataset that has 39260 sentences. This amount of data is very large compared to the current Indonesian AMR dataset.

Since AMR researches are still focused only on English, there are several challenges that needs to be addressed for Indonesian. First, there is no well-defined rule to describe relations among words, like English sentences with its Propbank framesets. Second, there is only one small dataset labeled AMR available for Indonesian news sentences [3]. These challenges needs to be addressed to create an AMR parsing system for Indonesian sentences.

Currently, there is only one work on AMR parsing for Indonesian. Severina & Khodra [3] developed rule-based AMR parser for multi-document summarization. AMR is used to capture concepts among news sentences from different sources. Similar concepts are merged to create a new AMR graph that contains important concepts from different sources. Since it uses manually-defined set of rules to parse sentence into AMR, this makes the AMR parsing system not scalable. They evaluated AMR parser using accuracy that only calculates the number of matching concepts between two AMRs, and obtained accuracy of 52.12%. The evaluation metrics should measure the correctness of the AMR using SMATCH, which measures the correctness of the concepts and the relation between the concepts [7].

Currently there is no work that employs machine learning approach to create an AMR parsing system for Indonesian sentences. In this paper, we propose a system that uses machine learning approach to create AMR parsing system. The system used *dependency parsing* features as its core features for the model.

In this paper, section II contains related works for Abstract Meaning Representation, and utilization of AMR. Section III contains proposed approach to the implemented system and components of the system. Section IV contains experiments that has been conducted and its discussion. Section V contains the conclusion from the result and future improvements for Indonesian AMR research.

## II. Related Work

### A. Abstract Meaning Representation

Abstract Meaning Representation (AMR) is defined by Banarescu et al. [1] to create a representation that can store many semantic information from a sentence. AMR is a rooted, directed, labeled, and acyclic graph that represents a sentence. Each AMR corresponds to the meaning of a sentence. There are several principles that is defined by Banarescu et al. [1] on AMR:

1. AMR is a graph that can be easily intepreted by humans and computer.
2. Sentences with similar meaning will be represented by the same AMR. For example, the sentence "I bought a book" and "The book is bought by me" will have the same AMR.
3. AMR uses the Propbank framesets, making it heavily reliant on English sentences.

Example of an AMR graph for the sentence *"Aku ingin makan kueku di Gedung Sabuga besok"* (I want to eat my cake in Sabuga building tomorrow) can be seen on Fig. 1. There are various features that is contained in an

AMR. Based on Fig. 1, semantic relations can be seen on the link between the word "*ingin*" (*want*) and "*aku*" (I) that shows the argument "ARG0" which means the word *"aku"* (I) is the actor that does the word *"ingin"* (want). AMR also supports coreferences. This feature can be seen on the link between the word *"kue"* (cake) and the word *"aku"* (I). *"aku"* (I) in that relation refers to the same *"aku"* (I) that acts as the actor of the word *"ingin"* (*"want")*. Named entites on AMR can be seen on the word "*Sabuga*" which has the attribute "*name*" indicating a named entiy of location.

There are some researches that uses AMR for various tasks. Severina & Khodra [3] used AMR for multi-document summarization to capture similar concepts from various news texts by creating AMR for every sentences in the news texts. Other application can be seen in paraphrase detection that is done by Issa et al. [2] that uses AMR to create the semantic representation among text and compares them whether it has the similar AMR or not.

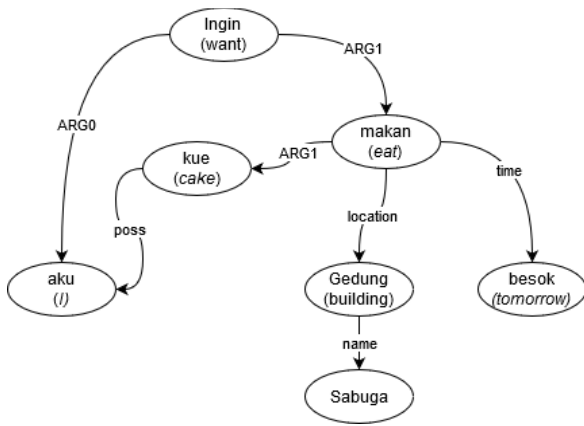

Fig. 1. AMR for "Aku ingin makan kueku di Gedung Sabuga besok"(*I want to eat my cake in Sabuga building tomorrow*)

*B. AMR Parsing*

A lot of research on AMR parsing has been done lately. However, most of the AMR parsing systems that are available focus on parsing AMR from English sentences. The current state-of-the-art AMR parsing system has been done by Zhang et al. [5]. Zhang et al. [5] was able to get the best SMATCH scores: 76.3% F1 on LDC2017T10 and 70.2% F1 on LDC2014T12.

Zhang et al. [5] separate the AMR parsing process into two tasks, node prediction and edge prediction. Zhang et al. [5] used *extended pointer generator network,* an improvement of *pointer generator network* by See et al. [6]. In addition to having the ability of doing the source copy, this model also has the ability to do a target copy, meaning that it can also point to the tokens that is used by the decoder. For the edge prediction task, Zhang et al. [5] used *biaffine classifier* for predicting the edges between the predicted words and used *bilinear classifier* for predicting the label for each edges. These two tasks, node prediction and edge prediction, are jointly trained. One of the main advantage of using this approach is this system requires no prealignment data, because it is a graph-based AMR parsing approach.

As for the research in Indonesian sentences, there has been only one research that uses AMR. Severina & Khodra [3] used AMR to conduct multi-docummment summarization with Indonesian news text. Severina & Khodra [3] used AMR to capture concepts between each sentences which are joined to collect the similar concepts among the sentences. This joined concepts are the features that are used to generate a summarization from text.

However, Severina & Khodra [3] used a rule-based approach to generate the AMR from the sentences. There are four steps in generating AMR from sentences from Indonesian sentences that is conducted by Severina & Khodra [3]. First, dependency parser captures ROOT word of the sentence. Second, words that are connected to the ROOT word are checked whether it is an active word or a passive word. Third, all other words are compared with the dictionary that contains specific label words. All of those information then is joined to create an AMR graph.

The rule-based approach employed by Severina & Khodra [3] created several limitations to the AMR produced from the system. First, node representation in AMR can be a phrase or multiple clauses, which means there are more than one concept depicted on one node. Second, the system limits the number of argument that can be detected to 3 only.

Comparing the AMR parsing approach between Severina & Khodra [3] and Zhang et al. [5], it is clear that Zhang et al. [5] produced a better representation of the AMR, as Zhang et al. [5] used the deep learning approach to create the AMR. This leads to a better ability for the model to generalize on each sentences. However, the *extended pointer generator network* that is used by Zhang et al. [5] needs a massive amount of data, which is also a limitation if we want to adapt it to AMR parsing for Indonesian. Compared with the *pretrained dependency parser* that is used by Severina & Khodra [3], this is way more feasible as it does not need massive amount of data to capture the dependency between words in a sentence.

### III. PROPOSED SOLUTION

Our system is designed based on the steps that are used by Zhang et al. [5] to create an AMR parsing sytem. Zhang et al. [5] used two phases: node prediction and edge prediction. Edge prediction contains two subtasks: edge prediction and label prediction. Therefore, there are three steps that were employed by Zhang et al. [5] to create an AMR parsing system. The comparation between the steps of the system of Zhang et al. [5] and the proposed system can be seen on TABLE I.

TABLE I. SYSTEM STEPS COMPARATION

| Research | Steps | | |
|---|---|---|---|
| | *Node Prediction* | *Edge Prediction* | *Label Prediction* |
| Zhang et al. [5] | Extended pointer generator network | Biaffine classifier | Bilinear classifier |
| Proposed System | Dependency parsing + pair filtering | | label classifier model |

The proposed system used the result of dependency parsing as features. These features create pairs of words with its dependency role that shows the connection between the two words. However, not all pairs of words produced by the dependency parsing are essential for the construction of the AMR graph, hence the pair filtering phase. Pair filtering aims to filter all of the pair of words that are not going to be used in the AMR. Label classifier model is a supervised model that predicts the correct AMR label given the features captured from the dependency parser. Detailed components of the system can be seen on Fig. 2.

Each of the individual components from Fig. 2 will be explained as the following.

*1) Dependency Parsing*

This component uses a *pretrained dependency parser* to capture the connection between the words that are contained in the sentence. These connections is the fundamental feature that will be used to create the AMR graph. We use *StanfordNLP[1]*, specifically the pretrained Indonesian model which was trained using the UD_Indonesian-GSD treebank. This pretrained dependency parser was able to get unlabeled attachment score (UAS) of 85.22 and labeled attachment score (LAS) of 79.17.

*2) Preprocessing*

This component aims to capture the features that are contained in a word. There are five preprocessing steps, namely punctuation removal, tokenization, lemmatization, named entity relation (NER) tagging, and part of speech (POS) tagging.

Punctuation removal is done using the Python string replacement function. This will remove all of the punctuation contained in a sentence. Tokenization will separate the sentence into words, creating an array of words contained in that sentence. We use the built in StanfordNLP function that automatically separates the sentence into words. Lemmatization will transform the words into its root form. We use Sastrawi to conduct the lemmatization. For NER tagging, we use pretrained Anago NER tagger that is trained using the Indonesian named entity data. For POS tagging, we use the NLTK POS tagger that uses the Indonesian corpus.

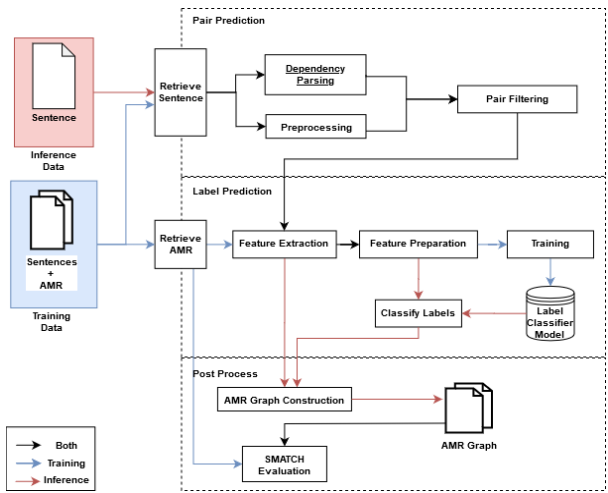

Fig. 2. Proposed System Architecture

*3) Pair Filtering*

This component is used to filter all of the unnecessary pair of words that will not be used in the AMR construction process. It will uses several rules to determine which pair of words that will be removed. Visualization of this process can be seen on Fig. 3.

Fig. 3. Pair filtering process

There are three rules that are used in this system, which are preposition rule, determiner rule, and subordinate conjunction rule. Preposition rule eliminates all pairs that contains prepositions. Example of words for this rule are "*di*" (in), "*ke*" (to), "*dari*" (from). Determiner rule eliminates all pairs that contains deteriminer words. Example of word for this rule is "*yang*" (which). Subordinate conjunction eliminates all pairs that contains subordinate conjunction words. Example of word for this rule is "*dengan*" (with).

*4) Feature Extraction*

This component aims to extract the features that are generated by the dependency parsing and preprocessing component. Edge pairs from the labeled AMR is matched with the corresponding pair features that were generated by the *dependency parser*. This component is divided into two steps: feature combining and pair matching.

Feature combining will combine all of the necessary features from the dependency parsing steps and the preprocessing steps. It will create a tabular data that contains all of the features that will be given to the AMR label classifier model. The combined feature will produce four feature categories, namely identifier feature, lexical features, syntactic features, and positional features. Detailed feature category for every combined feature is shown by TABLE II. These features will be used in the feature category experiment to determine which feature category combination gives the best performance.

TABLE II. DETAILED FEATURES

| No | Feature | Feature Category |
|---|---|---|
| 1 | Sentence ID | Identifier |
| 2 | Parent | Lexical |
| 3 | Child |  |
| 4 | Parent POS | Syntactic |
| 5 | Child POS |  |
| 6 | Parent NER |  |
| 7 | Child NER |  |
| 8 | Dependency role |  |
| 9 | Is Root |  |
| 10 | Parent position | Positional |
| 11 | Child position |  |

Pair matching is only conducted in the training process. This step is skipped in the inference process. Pair matching is conducted by iterating all of the features generated above for each AMR pairs in the dataset. If the *parent* and *child* feature matches with the *head* and *dependent* of the AMR pair, the label will be given to the feature. These labels will be used for the label classifier model. The visualization of pair matching can be seen on Fig. 4.

---
[1] https://stanfordnlp.github.io/stanfordnlp/

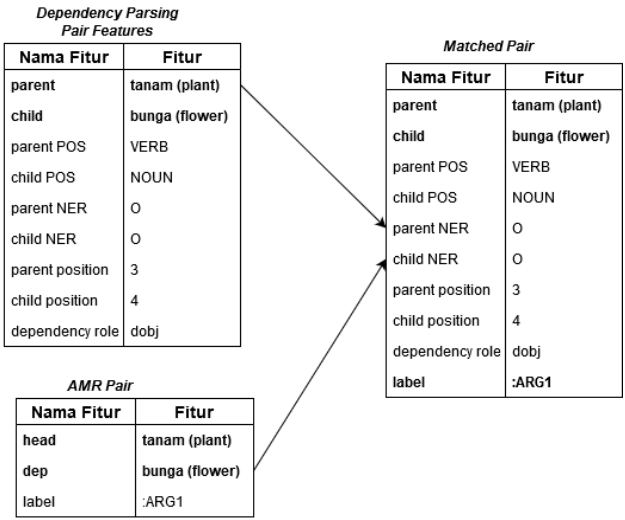

Fig. 4. Pair matching visualization

*5) Feature Preparation*

Feature preparation component will make sure that every attribute in the features generated in the feature extraction component can be used as the training or prediction data for the AMR label classifier model. There are two components in this step: word embedding and one hot encoding. Word embedding will change the word features to a vector representation. We use Gensim [2] Indonesian *word2vec* word embedding that has the length of 300 dimensions. All of the lexical features will be using this word embedding component. One hot encoding will change categorical features into its corresponding one hot encoded form. We use OneHotEncoder module provided by Scikit-learn to achieve this.

*6) Training and Label Classifying*

These components, training and label classifying, refers to the same component: the AMR label classifier model. This model predicts the AMR label given the feature provided from the feature preparation process. We use three supervised learning model: Decision Tree, XGBoost, and Feed Forward Neural Network. All of the model will be validated using the K-fold cross validation scheme. The best model is determined by the best F1 macro and will be saved for inference. Detailed parameters for the experiment will be explained in later section.

*7) AMR Graph Construction*

This component will construct an AMR graph given the AMR label that is predicted by the AMR label classifier model and the feature extracted from the feature extraction component. Every sentences will be transformed to an AMR graph that will be written to an external file. This file will be used for the SMATCH calculation, comparing the external file produced by the AMR graph construction component with the labeled gold AMR file.

---

[2] https://radimrehurek.com/gensim/models/word2vec.html

IV. DATA ANNOTATION

*A. Data Annotation*

To create an AMR parsing system that uses machine learning approach, a considerable amount of data is needed to make sure that the model can learn enough concept from the data. Therefore, data annotation is needed to support this goal. Because AMR annotation in Indonesian is a fairly new research, we limit the AMR labels (arguments) to six labels: *:ARG0, :ARG1, :name, :time, :location, :mod.* This limitation aims to simplify the labeling process and focus more on the quantity of the dataset.

We have collected a total of 1000 Indonesian simple sentences that are used for the training of the model, with the ratio of train:test is 70:30. We also reannotate the test data that were used by Severina & Khodra [3]. Reannotation is required because the test data that Severina & Khodra [3] used did not comply to the correct AMR specifications. The test data that Severina & Khodra [3] used still contained phrase as the node representation, as opposed to words. The detailed statistic of the dataset can be seen on TABLE III.

TABLE III. DATASET STATISTICS

| Data | Statistics | | |
|---|---|---|---|
| | Sentences Count | Node Count | Edge Count |
| Simple sentence – train | 700 | 3071 | 2371 |
| Simple sentence – test | 300 | 1395 | 1089 |
| b-salah-darat | 32 | 493 | 461 |
| c-gedung-roboh | 29 | 424 | 395 |
| d-indo-fuji | 27 | 555 | 528 |
| f-bunuh-diri | 23 | 321 | 298 |
| g-gempa-dieng | 19 | 286 | 267 |
| **Total** | **1130** | **6545** | **5409** |

V. EXPERIMENTS & DISCUSSION

*A. Pair Filtering Rule Experiment*

Our experiment aims to determine the best rule combination in the pair filtering phase. There are 7 rule combinations that are used that derived from the combination of the 3 pair filtering rules mentioned above: preposition, determiner, and subordinate conjunction (SC). The experiment is evaluated using *dependency pair* F1 metric. This metric is calculated by calculating the precision and the recall first. The precision can be calculated by counting the number of matching AMR pair and feature pair divided by the number of all pair features. Recall can be calculated by counting the number of matching AMR pair and feature pair divided by the number of AMR pair.

The result of this experiment can be seen on TABLE IV. Based on the result of the experiment, the rule combination of determiner, preposition, and subordinate conjunction used together yields the best F1 score. This best rule combination is used for the next experiment, which is the feature combination experiment.

The determiner, preposition, and subordinate conjunction combination yields the best result because they filter more words that is not necessary for the AMR graph. That is why the precision of the combination yields the highest value. This can happen because compared to the other rules, this rule will produce less words because it filter more words. As the number of prediction decreases, the precision will increase, assuming there are the same number of correct predictions for each combination.

TABLE IV. RULE COMBINATION EXPERIMENT RESULTS

| No | Rule Combination | | | Score | | |
|---|---|---|---|---|---|---|
| | Determiner | Preposition | SC | Precision | Recall | F1 Score |
| 0 | | | | 0.661 | 0.757 | 0.706 |
| 1 | ✓ | | | 0.751 | 0.737 | 0.744 |
| 2 | | ✓ | | 0.669 | 0.754 | 0.709 |
| 3 | | | ✓ | 0.727 | 0.754 | 0.740 |
| 4 | ✓ | ✓ | | 0.751 | 0.737 | 0.744 |
| 5 | ✓ | | ✓ | 0.688 | 0.737 | 0.711 |
| 6 | | ✓ | ✓ | 0.736 | 0.751 | 0.743 |
| 7 | ✓ | ✓ | ✓ | 0.761 | 0.734 | **0.747** |

*B. Feature Category Combination Experiment*

This experiment aims to determine the best feature category combination configuration that is used for the AMR label classifier model. The best feature combination is determined by calculating the F1 macro for each of the combination evaluated. This experiment will use a default decision tree classifier as the base model to determine the best feature combination. This experiment uses 5-fold cross validation scheme that is done with the *simple sentence – train* dataset. The result of this experiment can be seen in TABLE V.

TABLE V. FEATURE COMBINATION EXPERIMENT RESULTS

| No | Feature Category Combination | | | Nilai | |
|---|---|---|---|---|---|
| | Lexical | Syntactic | Positional | Accuracy | F1 Macro |
| 1 | ✓ | ✓ | ✓ | 0.843764 | 0.762202 |
| 2 | | ✓ | ✓ | 0.835649 | 0.714932 |
| 3 | ✓ | ✓ | | 0.841953 | **0.762866** |
| 4 | ✓ | | ✓ | 0.734734 | 0.707327 |
| 5 | | ✓ | | 0.83251 | 0.712700 |

Based on the result that is shown by TABLE V. the best feature category is achieved by using the lexical and syntactic feature category. This feature cateogry combination will be used for the next experiment, which is the algorithm and hyperparameter experiment. The result shows that positional feature category actually hurts the performance of the model. Compared to the first feature category combination which yields the higher accuracy but lower F1 macro which indicates the inability to detect labels from different classes.

*C. Algorithm & Hyperparameter Experiment*

This experiment aims to determine the best algorithm and their hyperparameter to create the AMR label classifier model. There are three algorithm that is used in this experiment, namely Decision Tree Classifier, XGBoost, and Feed Forward Neural Network. This experiment uses 5-fold cross validation scheme that is done with the *simple sentence – train* dataset. The detailed hyperparameter for each algorithm can be seen on TABLE VI. The best algorithm and hyperparameter combination is determined by comparing the F1 macro for each algorithm & hyperparameter combination.

The best algorithm and hyperparameter is achieved by using the XGBoost algorithm with *learning_rate = 0.1* and *max_depth = 8*. This algorithm achieved 0.904 accuracy and 0.880 F1 macro.

TABLE VI. ALGORITHM & HYPERPARAMETER EXPERIMENT DETAILS

| Model | Parameter | Nilai |
|---|---|---|
| Decision Tree Classifier | *max_depth* | 6, 7, 10, 12 |
| | *criterion* | *gini, entropy* |
| XGBoost | *learning_rate* | 0.05, 0.1, 0.2 |
| | *max_depth* | 5, 8, 10 |
| Feed Forward Neural Network | *units* | 50, 150, 300, 500 |
| | *layers* | 5,10,15 |

*D. AMR Parsing Evaluation*

This section will evaluate the proposed system using quantitative measure and a qualitative analysis. There are two models that are used to evaluate the AMR parsing system, our proposed system and the deep learning approach by Zhang et al. [5] (with some adaptation). We train both models with all of the simple sentence – train data. We use the simple sentence – test data and the reannotated data from Severina & Khodra [3] to conduct the evaluation. The proposed system employs the best pair filtering rule combination, feature category combination, and algorithm & hyperparameter combination that has been explained in the section before. The same parameters as Zhang et al. [5] are applied for the deep learning approach model. SMATCH score is used to compare the ability of the two models to generate the AMR given some sentences in a file. The result of the evaluation can be seen on TABLE VII.

TABLE VII. AMR PARSING SMATCH SCORES

| Data | Proposed System | Zhang et al. [5] |
|---|---|---|
| **Simple sentence** | 0.820 | 0.758 |
| **b-salah-darat** | 0.684 | 0.370 |
| **c-gedung-roboh** | 0.583 | 0.407 |
| **d-indo-fuji** | 0.677 | 0.318 |
| **f-bunuh-diri** | 0.587 | 0.437 |
| **g-gempa-dieng** | 0.672 | 0.406 |

Based on the results, the proposed system can perform better given the limited data. The deep learning approach by Zhang et al. [5] can also produce a decent score. Both of the models suffers from a more complex structured sentences. This could happen because of both of the models were trained only with the simple sentence dataset.

There are several errors that can be noticed in the proposed system AMR results, which is the semantic focus difference and lemmatization failures.

Semantic focus difference can be seen on the example as shown on TABLE VIII. The example shows two AMR for the "*Ibu menjahit baju dengan rapi*" (Mother sews the shirt neatly) sentence. Based on that example, the root of the AMR for the gold AMR should be the word "*jahit*" (sew) . Meanwhile, the proposed system outputs the prediction with the word "*ibu*" (mother) as the root of the AMR.

TABLE VIII. SEMANTIC FOCUS DIFFERENCE

| Sentence | Ibu menjahit baju dengan rapi<br>*Mother sews the shirt neatly* |
|---|---|
| Gold AMR | `(j / jahit [sew]`<br>`  :ARG0 (i / ibu [mother])`<br>`  :ARG1 (b / baju [shirt])`<br>`  :mod (r / rapi [neat]))` |
| System prediction | `(vv1 / ibu [mother]`<br>`  :mod (vv2 / jahit [sew]`<br>`    :ARG1 (vv3 / baju [shirt] )`<br>`    :mod (vv4 / rapi [neat]`<br>`      :mod (vv5 / dengan [with] ))))` |

The edges between the words in the AMR in the proposed system is produced by using the dependency parsing component. This means that component that constructs the edges is still very dependent on the result of the dependency parser. This means that the edge prediction component will have a better result if the dependency parser is improved. Therefore, advancement of the dependency parser or the decoupling between the proposed system with an existing dependency parser is needed to improve the proposed system.

The second error that can be noticed from the AMR result of the proposed system is the lemmatization failure. The lemmatization error example can be seen on TABLE IX.

TABLE IX. LEMMATIZATION FAILURE

| Sentence | Saya tertawa ketika melihat acara komedi di televisi<br>*I laugh when I see the comedy show on television* |
|---|---|
| Gold AMR | `(t1 / tawa [laugh]`<br>`  :ARG0 (s / saya [I] )`<br>`  :time (l / lihat [see]`<br>`    :ARG1 (a / acara [show]`<br>`      :mod (k / komedi [comedy] ))`<br>`  :location (t2 / televisi [television] )))` |
| System prediction | `(vv1 / tertawa [laughing]`<br>`  :ARG0 (vv2 / saya [I] )`<br>`  :mod (vv3 / lihat [see]`<br>`    :mod (vv4 / ketika [when] )`<br>`    :ARG1 (vv5 / acara [show]`<br>`      :mod(vv6 / komedi [comedy]))`<br>`  :location (vv7 / televisi [television] )))` |

From the example shown on TABLE IX. , the system prediction for the "*Saya tertawa ketika melihat acara komedi di televisi*" (I laugh when I see the comedy show on television) sentence failed to lemmatize the word "*tertawa*" (laughing). The correct word after lemmatization should be "*tawa*" (laugh) not "*tertawa*" (laughing). This means that the lemmatizer used for the system still has some errors that needs to be addressed. A better lemmatizer can improve the overall performance of this proposed system.

## VI. CONCLUSION AND FUTURE WORKS

We conclude that an AMR parsing system for Indonesian using machine learning approach can be built using three steps that is inspired by Zhang et al. [5] work. Those steps are *pair prediction*, *label prediction*, and the postprocess.

Our proposed system is able to produce decent result in a simple structured sentence, but still suffers in a more complex structured sentences. Our proposed system is able to reach reasonable SMATCH score of 0.820 for simple sentence – test data, 0.684 for b-salah-darat topic, 0.583 for c-gedung-roboh topic, 0.677 d-indo-fuji topic, 0.687 for f-bunuh-diri topic, and 0.672 for g-gempa-dieng topic.

Future improvements can be done in several aspects of this study. First, there should be a more focused research on the formalization of the arguments, like Propbank framesets for English sentences. Second, the addition of labeled data that contains more varying labels and more complex structured sentence data is needed to improve the model performance. Last, the proposed system needs to be decoupled from the dependency parser. This can be done by creating a model that can predict pairs without the reliance on a third party model.

ACKNOWLEDGMENT

We thank Zhang et al. [5] for the model repository that is useful for this study as reference.